\documentclass[conference]{IEEEtran}
\IEEEoverridecommandlockouts
\usepackage{cite}
\usepackage{amsmath,amssymb,amsfonts}
\usepackage{algorithmic}
\usepackage{graphicx}
\usepackage{url}
\usepackage{textcomp}
\usepackage{xcolor}

\usepackage{multirow}
\usepackage{booktabs}
\usepackage{placeins} 
\usepackage{dblfloatfix}  
\usepackage[font=footnotesize]{caption}
\usepackage{subcaption} 
\def\BibTeX{{\rm B\kern-.05em{\sc i\kern-.025em b}\kern-.08em
    T\kern-.1667em\lower.7ex\hbox{E}\kern-.125emX}}
\begin{document}

\title{The Value of Graph-based Encoding In NBA Salary Prediction\\

}

\author{\IEEEauthorblockN{1\textsuperscript{st} Junhao Su}
\IEEEauthorblockA{\textit{Department of Computer Science} \\
\textit{Brigham Young University}\\
Provo, UT, USA \\
sjh666@byu.edu}
\and
\IEEEauthorblockN{2\textsuperscript{nd} David Grimsman}
\IEEEauthorblockA{\textit{Department of Computer Science} \\
\textit{Brigham Young University}\\
Provo, UT, USA \\
grimsman@cs.byu.edu}
\and
\IEEEauthorblockN{3\textsuperscript{rd} Christopher Archibald}
\IEEEauthorblockA{\textit{Department of Computer Science} \\
\textit{Brigham Young University}\\
Provo, UT, USA \\
archibald@cs.byu.edu}

}

\maketitle

\begin{abstract}

Market valuations for professional athletes is a difficult problem, given the amount of variability in performance and location from year to year. In the National Basketball Association (NBA), a straightforward way to address this problem is to build a tabular data set and use supervised machine learning to predict a player's salary based on the player's performance in the previous year. For younger players, whose contracts are mostly built on draft position, this approach works well, however it can fail for veterans or those whose salaries are on the high tail of the distribution. In this paper, we show that building a knowledge graph with on and off court data, embedding that graph in a vector space, and including that vector in the tabular data allows the supervised learning to better understand the landscape of factors that affect salary. We compare several graph embedding algorithms and show that such a process is vital to NBA salary prediction.
\end{abstract}

\begin{IEEEkeywords}
Graph Representation Learning, Knowledge Graphs, Economic Valuation, Cold-Start Problem, Evaluation Methodology, Sports Analytics
\end{IEEEkeywords}
\section{INTRODUCTION}

In the high-stakes market of professional sports, player valuation transcends mere statistical calculation; it is a critical financial imperative. While the "Moneyball" era cemented the primacy of on-court metrics~\cite{scully1974pay, kahn1988racial}, modern salary formation is increasingly recognized as an "embedded" sociological process~\cite{granovetter1985economic}. A player's market value is driven not solely by individual statistics, but is profoundly shaped by \textit{relational capital}—such as agent negotiating leverage, team stylistic fit, and scarcity within the talent network~\cite{simmons2011mixing}. However, dominant tabular baselines (such as XGBoost) inherently treat players as isolated rows, collapsing rich structural dependencies into simplistic one-hot indicators~\cite{wu2018classification, xu2025enhanced}. This reduction inevitably sacrifices the multi-hop context that drives actual market inefficiencies.

Recent works apply Graph Neural Networks (GNNs) to model these complex interactions. Yet, a fundamental methodological question remains: does network topology offer a genuinely orthogonal signal, or is it just a redundant proxy for explicit metadata? Prior literature often conflates the two or suffers from temporal leakage~\cite{kaufman2012leakage, shchur2018pitfalls}, obscuring whether performance gains stem from actual structural reasoning or mere experimental artifacts. Without strict guardrails on information availability, the true utility of "structural capital"—distinct from explicit labels like Team or Agent IDs—remains fundamentally unproven.

To resolve this, we introduce a \textit{Matched-Information Evaluation Framework}. Moving beyond mere benchmark-chasing, we strictly isolate the independent predictive power of network topology. By benchmarking graph models against an explicit-ID Oracle,'' we reveal precisely when structural signals (who you know'') provide orthogonal utility over categorical metadata (``who you are'').

Our work moves beyond simple average accuracy to characterize the \textit{mechanistic role} of graph structure in economic valuation. We demonstrate three primary insights:
\begin{enumerate}
    \item \textbf{High-Fidelity Proxies:} Blind to explicit Team or Agent IDs, static structural embeddings recover a substantial fraction of the Oracle's predictive power. This proves topology alone successfully encodes latent institutional representations.
    
    \item \textbf{Orthogonal Risk Management \& Structural Maturity:} While tabular models dominate in average efficiency, graph models function as critical ``safety nets'' for tail-risk instances. We identify a clear phase transition in valuation logic: for established veterans, graph models rescue extreme outliers (reducing errors by $>\$10$M) by capturing the accumulated social capital that meritocratic tabular models miss. Conversely, emerging rookies exist in a ``structural vacuum'' where graph embeddings only introduce noise, cementing pure tabular formulations as the optimal paradigm for new entrants.
    
    \item \textbf{Signal Dilution:} Defying the ``more-is-better'' myth, dense heterogeneous graphs (such as V2-Full) fail to reliably outperform simpler topologies. This exposes a strict \textit{Quality over Quantity} dynamic, where specific affiliation edges far outweigh sheer volumes of noisy historical event logs.
\end{enumerate}
\section{Related Work}

\subsection{Economic Valuation and Relational Capital in Sports}

Quantitative analysis in professional sports has traditionally focused on identifying the statistical drivers of market value. Early econometric frameworks relied on linear regressions to map on-court productivity to compensation~\cite{scully1974pay,kahn1988racial,berri1999most}, while modern approaches increasingly utilize high-capacity non-linear estimators, such as XGBoost, to capture complex feature interactions~\cite{xu2025enhanced}. However, these purely meritocratic formulations treat players as isolated entities, largely ignoring the sociological concept of ``embeddedness''~\cite{granovetter1985economic}—the principle that economic outcomes are heavily dictated by relational structures. Standard tabular models struggle to differentiate the accumulated social capital of an established veteran from the raw statistical output of a rookie, missing the agent networks, franchise loyalty, and structural prestige that drive true market valuation~\cite{simmons2011mixing}.

\subsection{Graph Representation Learning and Structural Vulnerabilities}

To capture these multi-hop dependencies, recent research has adapted graph representation learning for sports analytics. Techniques range from static random-walk embeddings like Node2Vec~\cite{grover2016node2vec, guan2023nba2vec} and RotatE~\cite{sun2019rotate} to dynamic message-passing frameworks such as GraphSAGE~\cite{hamilton2017inductive} and R-GCN~\cite{schlichtkrull2018modeling}. However, applying graphs to economic valuation exposes known algorithmic vulnerabilities. Specifically, message-passing architectures are notoriously prone to \textit{oversmoothing}—which can dilute the unique features of highly connected nodes (e.g., superstars)—and suffer severe performance drops in \textit{cold-start} scenarios where new nodes (e.g., rookies) lack established edges. Our work bridges this gap by empirically linking these theoretical vulnerabilities to specific demographic cohorts within the sports labor market.

\section{METHODOLOGY}

\subsection{Data Sources and Baselines}
We formulate the valuation task across five NBA seasons (2020--21 through 2024--25), targeting log-annual salary, $y=\log(1+\text{salary})$. To capture comprehensive economic context, we seasonally align heterogeneous data: on-court stats(\textit{NBA.com}), team valuations (\textit{Forbes}), agencies (\textit{RealGM}), awards (\textit{Basketball-Reference}), and injury logs (\textit{Kaggle}). A reproducibility package is available in the GitHub repository referenced in the Acknowledgments.

The data was aggregated into a single table, where each row corresponds to a player $p$ and a season $s$, i.e., one row would correspond to $p = $ Nikola Jokic and $s = $2022-2023. Each row $(p, s)$ contains on-court statistics $x_{p,s}$ and career controls $c_{p,s}$ (draft position, age, etc.). The model we will refer to as \textit{Weak Baseline} is a model that predicts next year's salary based solely on $x_{p, s}$ and $c_{p,s}$.

In addition to this data, the model we refer to as \textit{Strong Baseline} is also a function of $m_{p,s}$, which are labels that refer to the team and agent for player $p$ during season $s$. The performance of the Strong Baseline model to the Weak Baseline model gives an indication as to how valuable the information $m_{p,s}$ is to these predictions (see Table \ref{tab:main_results}). 


\subsection{Data and Knowledge Graph Construction}

\begin{figure}[t]
    \centering    \includegraphics[width=0.9\columnwidth]{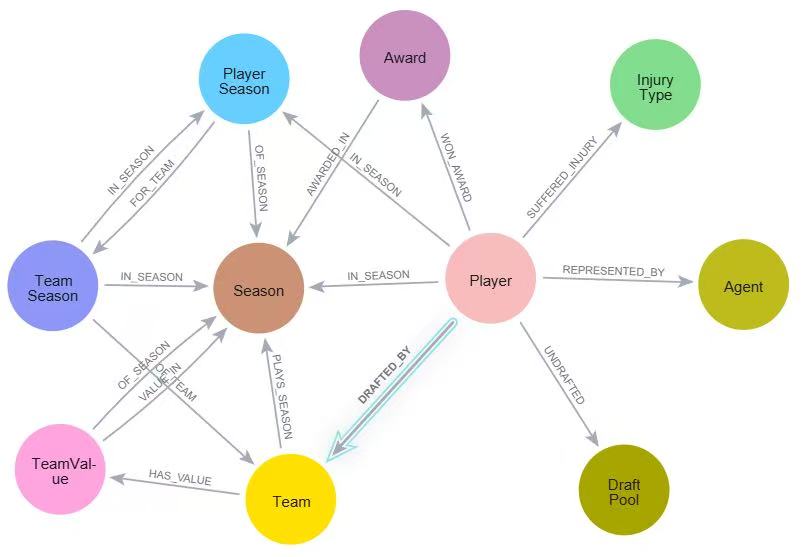} 
    \caption{\textit{Schema of the Heterogeneous NBA Knowledge Graph.} The graph connects \texttt{PlayerSeason} anchor nodes (center) to diverse entities including \texttt{Team}, \texttt{Agent}, \texttt{Award}, and \texttt{Injury}. Temporal edges (e.g., \textsc{Won\_Previously}, \textsc{Has\_Injury\_History}) are strictly masked by the admissibility function $A(e,s)$ to prevent look-ahead bias.}
    \label{fig:kg_schema}
\end{figure}

In order to understand the importance of structural data in this prediction problem, we construct a knowledge graph using $m_{p,s}$. From that graph, we embed each player-season node in a vector space, resulting in vector $z_{p, s}$. A model is then trained on $x_{p,s}$, $c_{p,s}$ and $z_{p,s}$ - the labels $m_{p,s}$ are withheld.

Indexing seasons by start year $\mathcal{S}=\{2020,\dots,2024\}$, we model the off-court context for player $p$ in season $s$ as a typed knowledge graph $\mathcal{G}=(\mathcal{V},\mathcal{E})$ (Figure~\ref{fig:kg_schema}). The entity set $\mathcal{V}$ comprises career-level nodes (\texttt{Player}, \texttt{Team}, \texttt{Agent}) and a \texttt{PlayerSeason} prediction anchor $v_{p,s}$, which edges connect to institutional and event entities. To prevent trivial target leakage, edges strictly encode relational existence, excluding all historical salary or contract values.

\subsection{Graph Embedding Methods}

In order to understand how graph embeddings affect the model predictions, we compare the results from several different graph embedding methods.

\subsubsection{Tabular Baselines \& Static Embeddings}

The \textit{Weak Baseline} relies solely on performance and temporal metrics, whereas the \textit{Strong Baseline (Explicit-ID Upper Bound)} memorizes market biases via explicit Team/Agent IDs. To extract structural signals without explicit labels, We also include two unsupervised static embedding baselines.
\textit{Node2Vec} learns node vectors from biased random walks that preserve local neighborhood structure \cite{grover2016node2vec},
while \textit{RotatE} represents relations as rotations in complex space to capture relational patterns \cite{sun2019rotate}, operating globally and transductively, these are interpreted separately from our inductive claims.

\subsubsection{Graph Neural Networks}

Crucially, all GNNs are strictly blind to explicit IDs, inferring context purely via connectivity.

First, \textit{V1 (Static GNN)} applies \textit{GraphSAGE}~\cite{hamilton2017inductive} over overarching \textit{Player} entities, injecting safe temporal offsets into the prediction head without altering the static topology.

Second, \textit{V2-Base (Dynamic GNN)} centers on dynamic \textit{PlayerSeason} nodes. We evaluate it \textit{Transductively} (V2-Trans, retaining unlabeled test nodes for global awareness) and \textit{Inductively} (V2-Ind, strictly masking test nodes during training to evaluate inference on emergent structures).

Finally, \textit{V2-Full (Heterogeneous Dynamic GNN)} relaxes the matched-information constraint, using \textit{R-GCN}~\cite{schlichtkrull2018modeling} to inject dense semantic events (Awards, Injuries). This tests if explicit institutional reality enhances or dilutes the core signal. We contrast a Single-Graph (\textit{V2-Full SG}, testing mere event existence) with a Multi-Graph variant (\textit{V2-Full MG}, retaining parallel edges to capture cumulative intensity as a proxy for prestige).

\subsection{Evaluation Protocols}
We adopt a strict forecasting setup. Models are trained on seasons 2020--2022, tuned on 2023, and evaluated on 2024 (the 2024--25 season). To guarantee comparability while mitigating selection effects, we define a shared test subset consisting of 668 player-season instances in 2024. This intersection is defined with respect to core modalities shared by all models; the sparse event modalities (Awards, Injuries) introduced in V2-Full are not required for inclusion.

\paragraph{Tri-State Rescue and Misguidance Protocol}
While average error metrics measure global fit, they obscure the localized impact of topological signals on outliers. We introduce a discrete evaluation protocol operating in the original dollar space. We first isolate a pool of ``Eligible Outliers'' where the baseline model exhibits significant deviation ($|Y-\hat{Y}_{\text{base}}|>\tau_{err}$). We set $\tau_{err}$ as the 75th percentile of baseline absolute residuals on the combined training and validation seasons (yielding $\$1.0\text{M}$ for the Weak Baseline and $\$0.7\text{M}$ for the Strong Baseline), and hold these thresholds fixed for all test analyses. For these instances, we define the correction margin as 
\begin{equation*}
    \Delta E=|Y-\hat{Y}_{\text{base}}|-|Y-\hat{Y}_{\text{graph}}|
\end{equation*}
and categorize the graph model's impact into three states: \textit{Successful Rescue} ($\Delta E>\$0.5\text{M}$, representing a substantial, non-trivial contract adjustment) where the graph embedding reduces error; \textit{Neutral Margin} ($-\$0.5\text{M}\le\Delta E\le\$0.5\text{M}$), where the graph embedding and baseline models predict roughly the same outcome; and \textit{Structural Misguidance} ($\Delta E<-\$0.5\text{M}$) where the graph embedding performs worse than baseline by more than the correction margin (see Figure~\ref{fig:tri_state}). This symmetric formulation prevents survivorship bias, transparently reporting both the \textit{Rescue Rate} and the \textit{Misguidance Rate} (the ``generalization tax'' of our structural modeling).

\paragraph{Qualitative Analysis and Example Selection}
To identify representative cases without selection bias, we sample examples strictly from the pool of successful rescues and misguidance failures. We prioritize coverage across four error modes: Underrated versus Overrated, crossed with Precision versus Overshoot. For each architecture, we algorithmically select the single instance with the largest absolute correction margin ($|\Delta E|$) within each category. If outcomes do not populate at least three quadrants, we fill remaining slots by descending absolute margin, ensuring a minimum of three examples per model. This deterministic protocol guarantees cited cases reflect empirical manifestations rather than cherry-picked anecdotes, see Table~\ref{tab:model_cases} for this analysis.

\paragraph{Quantitative Feature Profiling of Outliers}
While qualitative examples illustrate individual corrections, we introduce a systemic profiling protocol to identify the shared characteristics driving each model's rescue or misguidance behavior. Crucially, this profiling is strictly an \textit{ex post} characterization: these tabular features are not provided to the graph prediction head as identifiers, but serve only to describe the rescued versus misled cohorts. 

For a given baseline context, we partition the Eligible Outliers into a \textit{Rescue Set} $\mathcal{R}$ ($\Delta E > \$0.5\text{M}$) and a \textit{Misguidance Set} $\mathcal{M}$ ($\Delta E < -\$0.5\text{M}$). We profile these sets using the common tabular features (player demographics, career controls, and on-court statistics) available on the 2024 intersection split. Because sports performance metrics frequently violate normality assumptions, we evaluate statistical significance using the non-parametric Mann-Whitney U test, complemented by Cliff's Delta ($\delta$) to quantify the effect size. 

For a specific feature, let $x_i \in \mathcal{R}$ and $x_j \in \mathcal{M}$ represent the feature values for instances in the two cohorts, with sizes $n_R$ and $n_M$ respectively. Cliff's Delta is defined as the mean of the signum of all pairwise differences:
$$ \delta = \frac{1}{n_R n_M} \sum_{i=1}^{n_R} \sum_{j=1}^{n_M} \text{sgn}(x_i - x_j) $$
where $\text{sgn}(\cdot)$ is the signum function. The metric ranges from $-1$ to $1$, with the sign explicitly indicating whether the Rescue cohort systematically exhibits larger (positive $\delta$) or smaller (negative $\delta$) values than the Misguidance cohort. 

To extract the most distinguishing characteristics, we apply a dual-threshold filtering mechanism. For each graph architecture independently, we retain features satisfying both statistical significance ($p \le 0.10$) and a meaningful effect size ($|\delta| \ge 0.25$). This additional effect-size constraint reduces sensitivity to multiple comparisons and small-sample artifacts. Qualifying traits are then ranked by $|\delta|$ in descending order, with the top-$K$ ($K=8$) features reported as the model's \textit{top traits}. These ranked profiles summarize stable cohort differences consistent with hypothesized structural behaviors (e.g., neighborhood regularization vs. structural lag). See Table~\ref{tab:top_traits} for this analysis.

\subsection{Implementation Details}
All experiments were conducted using PyTorch and Scikit-Learn, averaged over 5 random seeds. Tabular baselines utilized Random Forest (500 estimators) and XGBoost, with the latter tuned using a learning rate of 0.03 and early stopping on the validation set. Categorical features in the Strong Baseline were processed via ordinal encoding fitted strictly on the training set to prevent data leakage, and continuous meta-features were imputed using training medians. Graph embedding models utilized dimension $d=64$ (Node2Vec) and $d=128$ (RotatE in $\mathbb{C}^{d}$). GNNs utilized a 2-layer architecture with hidden dimension $d=64$, trained with AdamW ($\text{LR}=3e^{-3}$) for 200 epochs. Fusion concatenated the pre-trained structural embedding $z_{p,s}$ with tabular features prior to the downstream regressor.
\section{RESULTS}

\begin{figure*}[!t]

    \centering

    \includegraphics[width=0.60\textwidth, trim=15 5 15 5, clip]{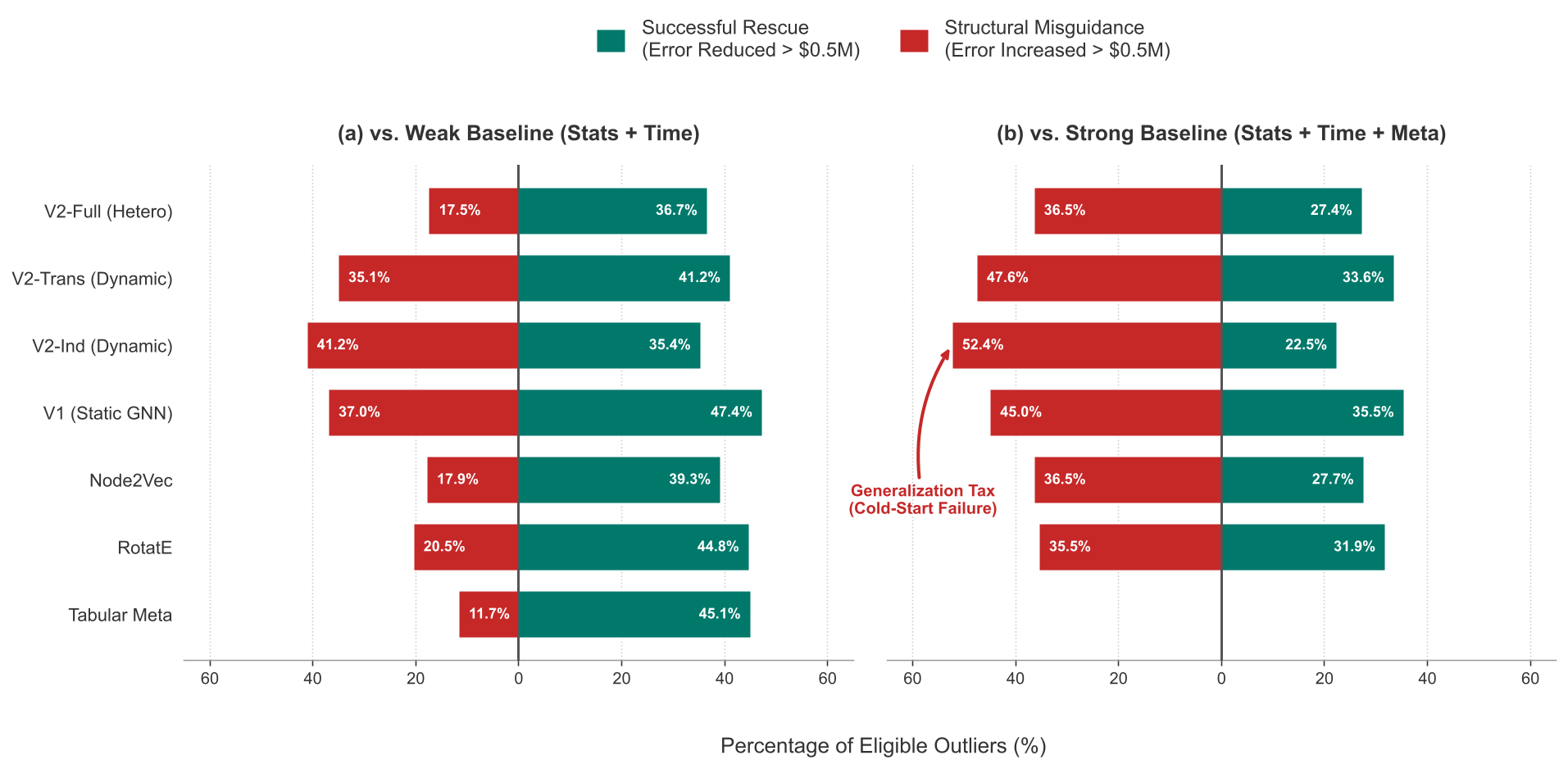}

    \caption{Tri-State Evaluation on Eligible Outliers. (a) vs. Weak Baseline: Static embeddings provide a favorable rescue--misguidance trade-off. (b) vs. Strong Baseline: Dynamic architectures incur a ``Generalization Tax,'' reflecting sensitivity to historical networks.}

    \label{fig:tri_state}

\end{figure*}

We evaluate the proposed framework across four dimensions: (1) global predictive accuracy, including a strict cold-start analysis; (2) localized risk mitigation via the Tri-State protocol; (3) quantitative feature profiling of rescued cohorts; and (4) structural ablation checks.

\subsection{Global Predictive Performance and the Cold-Start Vacuum}

Table~\ref{tab:main_results} summarizes the performance on the 2024 forecasting split. The results reveal a distinct interaction between structural representations, downstream regressor capacity, and the historical depth of the evaluated entities.

When utilizing Random Forest on the full test set, static graph embeddings demonstrate significant orthogonal value. Specifically, \textit{RotatE + Stats} reduces the log-salary RMSE to 0.654 compared to the \textit{Weak Baseline} (0.691). This confirms that topological connectivity functions as a high-fidelity proxy, effectively recovering latent entity representations without explicit identifiers. However, introducing high-capacity gradient boosting (XGBoost) and explicit metadata (\textit{Strong Baseline}) establishes a performance ceiling (RMSE 0.598). Under these strict conditions, dynamic GNNs (V1, V2) yield diminishing returns in average global error, suggesting that explicit tabular IDs efficiently absorb the macroeconomic premiums that graphs otherwise attempt to infer.

Crucially, the global metrics obscure a sharp divergence in model behavior when predicting salaries for new entrants to the league. As detailed in the Cold-Start column of Table~\ref{tab:main_results}, when evaluating exclusively on a ``Cold-Start'' subset (players appearing in the 2024 test set who were absent from the training seasons), the Strong Baseline (+Meta IDs) maintains strong predictive viability ($R^2 \approx 0.53$). In contrast, graph models—particularly inductive architectures (V2-Ind)—experience severe performance degradation, completely collapsing into negative predictability ($R^2 \approx -0.31$). This indicates that in a structural vacuum, where entities lack dense historical interaction edges, graph embeddings introduce noise that disrupts the otherwise stable rule-based inference derived from tabular demographics (like age and draft pick). Consequently, pure tabular formulations remain the optimal predictive paradigm for evaluating emerging rookies.

\begin{table}[!t]
    \centering
    \caption{Global Predictive Performance. Models ranked by Random Forest (RF) RMSE. \textit{Baseline (+Meta)} represents the Explicit-ID Upper Bound. The Cold-Start column exclusively evaluates new entrants, demonstrating the collapse of graph models in structural vacuums ($\downarrow$: lower is better, $\uparrow$: higher is better).}
    \label{tab:main_results}
    \scriptsize
    \setlength{\tabcolsep}{4pt}
    \renewcommand{\arraystretch}{1.1}
    \begin{tabular}{lccccc}
        \toprule
        \multirow{2}{*}{\textbf{Feature Set}} & \multicolumn{2}{c}{\textbf{Global RMSE} $\downarrow$} & \multicolumn{2}{c}{\textbf{Global $R^2$} $\uparrow$} & \textbf{Cold-Start} $\uparrow$ \\
        \cmidrule(lr){2-3} \cmidrule(lr){4-5} \cmidrule(lr){6-6}
        & \textbf{RF}  & \textbf{XGB}  & \textbf{RF}  & \textbf{XGB} & \textbf{$R^2$ (XGB)} \\
        \midrule
        Baseline (+Meta)      & \textit{0.618} & \textit{0.598} & \textit{0.717} & \textit{0.735} & \textit{0.535} \\
        RotatE + Stats        & 0.654 & 0.668 & 0.683 & 0.670 & -0.014 \\
        Node2Vec + Stats      & 0.659 & 0.665 & 0.678 & 0.672 & -0.004 \\
        V2-Full MG + Stats    & 0.687 & 0.695 & 0.650 & 0.642 & -0.047 \\
        V2-Full SG + Stats    & 0.689 & 0.702 & 0.648 & 0.635 & -0.047 \\
        Baseline (Stats+Time) & 0.691 & 0.697 & 0.646 & 0.640 & \textit{0.069} \\
        \midrule
        V2-Ind + Stats        & 0.735 & 0.787 & 0.600 & 0.541 & -0.306 \\
        V2-Trans + Stats      & 0.739 & 0.771 & 0.595 & 0.560 & -0.292 \\
        V1 + Stats            & 0.758 & 0.747 & 0.575 & 0.587 & -0.324 \\
        \bottomrule
    \end{tabular}
\end{table}

\subsection{Tri-State Evaluation: Risk-Reward and the Crowding-Out Effect}

Average error metrics obscure the localized, high-leverage impacts of topology on market outliers. Applying the Tri-State Evaluation Protocol to the ``Eligible Outliers'' subset (Figure~\ref{fig:tri_state}) exposes a clear risk-reward trade-off and highlights the crowding-out effect of explicit metadata.

Against the \textit{Weak Baseline}, static embeddings act as effective, low-risk proxies. RotatE achieves the most favorable trade-off (44.8\% rescue vs. 20.5\% misguidance). Without access to explicit labels, graph topology efficiently corrects severe pricing deviations without inducing widespread collateral errors.

Conversely, against the \textit{Strong Baseline}, we observe a pronounced crowding-out effect. As explicit tabular IDs efficiently encode entity-level premiums, the marginal utility of structural representations diminishes, causing rescue rates to drop and misguidance rates to rise across all models. Furthermore, dynamic message-passing architectures incur a notable \textit{Generalization Tax} in this information-rich environment. Specifically, V2-Ind's misguidance rate increases to 52.4\%. This underscores a vulnerability to structural lag and feature interference, where inductive models regularize toward historical local neighborhoods rather than yielding to concurrent, highly predictive explicit features.

\subsection{Deconstructing the Structural Proxy: Pedigree vs. Tenure}

To systematically explain \textit{why} graph models correct specific cohorts, we apply the dual-threshold profiling protocol ($p \le 0.10, |\delta| \ge 0.25$) to the empirical outlier distributions. Table~\ref{tab:top_traits} reveals that the graph representation fundamentally shifts its proxy role based on the available tabular context, transitioning from institutional pedigree to accumulated tenure.

\textbf{Proxy for Pedigree:} Under the \textit{Weak Baseline}, graph models exhibit a strong preference for draft capital. For instance, Node2Vec yields a Cliff's $\delta$ of -0.34 for \texttt{round\_pick}, indicating that successful rescues are systematically concentrated among high-draft-pick players (lower numerical values). When a purely performance-based tabular model underestimates a young lottery pick due to modest short-term efficiency, the graph topology corrects this by implicitly reconstructing the player's draft pedigree and market expectation through connectivity alone.

\textbf{Proxy for Tenure:} When explicit identifiers are introduced (\textit{Strong Baseline}), the graph models' rescue demographic shifts entirely. The defining traits transition to \texttt{age\_now} (RotatE $\delta=0.38$) and \texttt{draft\_year} ($\delta=-0.41$). This demonstrates that once team and agent identities are known to the tabular baseline, the primary orthogonal value of the graph structure lies in quantifying accumulated social capital and long-term league tenure. The network topology serves as a valuation safety net for aging veterans, capturing the structural prestige that protects their contracts even as their short-term athletic performance metrics decline. Therefore, this empirically establishes that graph architectures are uniquely suited for modeling the valuations of established veterans.

\begin{table}[!t]
    \centering
    \caption{Top Profiling Traits separating Rescue vs. Misguidance Cohorts ($p \le 0.10, |\delta| \ge 0.25$). Positive $\delta$ indicates the trait is significantly higher in successfully rescued players; negative indicates higher in misguided players.}
    \label{tab:top_traits}
    \resizebox{\columnwidth}{!}{%
        \renewcommand{\arraystretch}{1.1}
        \begin{tabular}{lllcc}
            \toprule
            \textbf{Baseline Context} & \textbf{Graph Model} & \textbf{Top Feature} & \textbf{Cliff's $\delta$} & \textbf{$p$-value} \\
            \midrule
            \multirow{5}{*}{\begin{tabular}[c]{@{}l@{}}\textit{Weak} \\ \textit{(Stats+Time)}\end{tabular}} 
            & Node2Vec & \texttt{round\_pick} & -0.34 & $<0.001$ \\
            & Node2Vec & \texttt{overall\_pick} & -0.31 & $0.001$ \\
            & RotatE & \texttt{round\_pick} & -0.26 & $0.003$ \\
            & V2-Ind & \texttt{TS\%\_calc} & -0.26 & $<0.001$ \\
            & V2-Trans & \texttt{overall\_pick} & -0.28 & $<0.001$ \\
            \midrule
            \multirow{6}{*}{\begin{tabular}[c]{@{}l@{}}\textit{Strong} \\ \textit{(+Meta)}\end{tabular}} 
            & RotatE & \texttt{draft\_year} & -0.41 & $<0.001$ \\
            & RotatE & \texttt{age\_now} & 0.38 & $<0.001$ \\
            & Node2Vec & \texttt{draft\_year} & -0.35 & $<0.001$ \\
            & V2-Full & \texttt{age\_now} & 0.34 & $<0.001$ \\
            & V2-Trans & \texttt{age\_now} & 0.31 & $<0.001$ \\
            & V2-Trans & \texttt{years\_since\_draft} & 0.26 & $0.001$ \\
            \bottomrule
        \end{tabular}%
    }
\end{table}

\subsection{Micro-Level Corrections and Structural Validity}

Table~\ref{tab:micro} provides deterministically selected case studies (largest $|\Delta E|$) that ground these statistical traits in empirical reality. For example, RotatE reduces the absolute error for Fred VanVleet by over \$9M across both baselines, acting as a proxy for his latent structural prestige. Conversely, it misguides the valuation of Chris Paul by anchoring too heavily on historical hubs rather than recent performance degradation, aligning with the \texttt{age\_now} risks identified in the profiling.


\begin{table}[!t]

    \centering

    \caption{Deterministically selected micro-level corrections for \textit{RotatE}. $\Delta E = |Y-\hat{Y}_{\text{base}}| - |Y-\hat{Y}_{\text{graph}}|$.}

    \label{tab:micro}

    \footnotesize

    \setlength{\tabcolsep}{4pt}

    \renewcommand{\arraystretch}{1.1}

    \begin{tabular}{llrrr}

        \toprule

        \textbf{Baseline} & \textbf{Player (Case)} & $\hat{Y}_{\text{base}}$ & $\hat{Y}_{\text{graph}}$ & $\Delta E$ \\

        \midrule

        \multirow{2}{*}{\textit{Weak}} 

        & F. VanVleet (Rescue) & \$7.47M & \$17.4M & \textbf{+\$9.97M} \\

        & Chris Paul (Misguide) & \$13.7M & \$23.1M & \textbf{-\$9.43M} \\

        \midrule

        \multirow{2}{*}{\textit{Strong}} 

        & F. VanVleet (Rescue) & \$8.08M & \$17.4M & \textbf{+\$9.37M} \\

        & Chris Paul (Misguide) & \$14.7M & \$23.1M & \textbf{-\$8.36M} \\

        \bottomrule

    \end{tabular}

\end{table}

\section{DISCUSSION}

\begin{table*}[t]
\centering
\caption{Model-centric representative cases selected deterministically from the eligible-outlier pool (2024 season). We report the maximum rescue ($\Delta E>0$) and misguidance ($\Delta E<0$) cases, with $\Delta E = |Y-\hat{Y}_{\text{base}}|-|Y-\hat{Y}_{\text{graph}}|$. All values are in \$M.}
\label{tab:model_cases}
\resizebox{0.85\textwidth}{!}{%
\renewcommand{\arraystretch}{0.95} 
\begin{tabular}{l|ll|ll}
\toprule
\multirow{2}{*}{\textbf{Graph Model}} 
& \multicolumn{2}{c|}{\textbf{Weak Baseline (Stats + Time)}} 
& \multicolumn{2}{c}{\textbf{Strong Baseline (+ Meta IDs)}} \\
\cmidrule{2-5}
& \textbf{Max Rescue ($\Delta E$)} & \textbf{Max Misguide ($\Delta E$)} & \textbf{Max Rescue ($\Delta E$)} & \textbf{Max Misguide ($\Delta E$)} \\
\midrule
\textbf{RotatE}   & VanVleet (+10.0) & Paul (-9.4)      & VanVleet (+9.4)  & Paul (-8.4) \\
\textbf{Node2Vec} & Oubre Jr. (+5.2) & Quickley (-5.1)  & White (+4.6)     & Reid (-3.6) \\
\textbf{V1}       & Middleton (+12.1)& Bane (-22.2)     & Middleton (+11.7)& Bane (-20.7) \\
\textbf{V2-Ind}   & Beasley (+7.5)   & Antetokounmpo (-11.5) & Beasley (+7.5) & Antetokounmpo (-12.2) \\
\textbf{V2-Trans} & Zubac (+10.0)    & Bane (-20.6)     & Zubac (+9.8)     & Bane (-19.1) \\
\textbf{V2-Full}  & Edwards (+2.8)   & LaVine (-3.4)    & Bogdanovi\'{c} (+2.7) & Reid (-4.4) \\
\bottomrule
\end{tabular}%
}
\end{table*}
Empirically, player valuation exhibits a stark tenure-based divide. While tabular models minimize average error, graph architectures specifically correct tail-risk outliers. We propose a ``Structural Maturity Hypothesis'': graph models capture veterans' accumulated social capital, whereas tabular baselines better reflect rookies' rigid, rule-based pricing. We also delineate graph failure boundaries and structural over-reliance risks.

\subsection{The Structural Maturity Hypothesis: Veterans vs. Rookies}

\textbf{The Veteran Premium: Graph as Social Capital.}
For established veterans, stable network affiliations (e.g., elite agencies, franchise loyalty) act as a valuation safety net. Tabular models, over-reliant on recent stats, systematically underestimate veterans during performance dips. Graph models correct this by capturing latent social capital (Table~\ref{tab:model_cases}). For instance, \textit{RotatE} improves Fred VanVleet's prediction by +\$10.0M; despite fluctuating stats, his elite agent network signals top-tier bargaining power. Similarly, \textit{V1} corrects Khris Middleton by +\$12.1M. While stats-driven baselines heavily penalize his recent injuries, the graph recognizes his decade-long franchise cornerstone status. Ultimately, graphs quantify implicit capital missed by standard metrics.

\textbf{The Rookie Vacuum: Tabular Superiority.}
Conversely, in Cold-Start evaluations for rookies, graph models collapse (e.g., $R^2 \approx -0.31$), while matched tabular baselines(Stats + Time) remain predictive ($R^2 \approx 0.07$). Rookie valuation is a \textit{rule-based} process driven by demographics (e.g., age, draft pick), not networks. Lacking deep teammate or commercial ties, rookies exist in a ``structural vacuum.'' Aggregating sparse neighborhoods merely injects noise, masking highly predictive explicit signals. Thus, for players lacking structural maturity, traditional tabular models remain optimal.

\subsection{Boundary Conditions and Structural Risks}

While successfully capturing prestige, network topology can also mislead predictions (Table~\ref{tab:model_cases}).

\textbf{Legacy Hangover.}
The mechanism saving veterans causes severe overestimation when historical status outpaces physical reality. For example, \textit{RotatE} overestimates Chris Paul by \$9.4M, anchoring on historical prestige while ignoring age-driven market devaluation.

\textbf{Structural Inertia.}
Dynamic message-passing (\textit{V1, V2-Trans}) struggles with ``breakout'' players. Desmond Bane (-\$22.2M error under V1) saw a rapid value spike, but neighborhood aggregation pulled his representation toward lower-salaried teammates, dragging predictions down.

\textbf{Hub Oversmoothing.}
Inductive GNNs (\textit{V2-Ind}) oversmooth high-degree superstars. Giannis Antetokounmpo (-\$11.5M error) is mathematically averaged with lower-salaried teammates during message passing, diluting his unique premium and highlighting a critical limitation of standard aggregation in star-driven markets.

\subsection{Implicit Proxies and the Tabular Ceiling}

These cases contextualize our macro-level findings. Given explicit IDs (\textit{Strong Baseline}), XGBoost easily memorizes entity premiums, creating a ceiling that makes graph embeddings redundant for average error reduction. However, under strict matched-information constraints (\textit{Weak Baseline}), static embeddings (\textit{RotatE}) act as powerful \textit{Structural Proxies}. By reconstructing missing institutional context solely through network connectivity, they prove topology carries robust economic signals without explicit labels.
\section{CONCLUSION}

This study isolates the independent predictive value of relational structure in economic valuation. Under strict anti-leakage constraints, we establish when topological networks ("who you know") provide orthogonal utility to explicit metadata ("who you are"), yielding three core insights:

High-Fidelity Proxies: Blind to explicit labels, static embeddings bridge much of the predictive gap to explicit-ID models, confirming they encode genuine relational patterns rather than covertly memorizing labels.

The Structural Maturity Dichotomy: Valuation logic exhibits a stark phase transition. For veterans, graphs capture accumulated social capital, hedging against stats-driven undervaluation. Conversely, for rookies lacking network histories, graph embeddings merely inject noise. Thus, rookie pricing remains a rule-based tabular problem, while veteran valuation is deeply relational.

Signal Dilution: Defying "more-is-better" assumptions, dense heterogeneous graphs failed to consistently outperform simpler topologies. The specificity of affiliation edges far outweighs the raw volume of historical event logs, enforcing a strict "quality over quantity" dynamic.

Ultimately, optimal valuation systems must operate as maturity-aware hybrids—defaulting to tabular baselines for rookies while activating graph modules for veterans. Future architectures should explore Graph Structure Learning (GSL) to prune topological noise and Macroeconomic Context Nodes to anchor pricing in league-wide financial realities.
\section*{Acknowledgments}
 We thank Dr. Chengkai Li for his invaluable support. AI assisted solely with language polishing; all analyses remain our own.Code is available at https://github.com/SergeSu111/nbasalaryKGNewst.
\bibliographystyle{IEEEtran}  
\bibliography{refs}           

@article{wu2018classification,
  title={Classification of NBA salaries through player statistics},
  author={Wu, William and Feng, Kevin and Li, Raymond and Sengupta, Kunal and Cheng, Austin},
  journal={Sports Analytics Group at Berkeley, \url{https://sportsanalytics.berkeley.edu/projects/nba-salaries-stats.pdf} [19.09.2020]},
  year={2018}
}

@article{simmons2011mixing,
  title={Mixing the princes and the paupers: Pay and performance in the National Basketball Association},
  author={Simmons, Rob and Berri, David J},
  journal={Labour Economics},
  volume={18},
  number={3},
  pages={381--388},
  year={2011},
  publisher={Elsevier}
}

@article{granovetter1985economic,
  title={Economic action and social structure: The problem of embeddedness},
  author={Granovetter, Mark},
  journal={American journal of sociology},
  volume={91},
  number={3},
  pages={481--510},
  year={1985},
  publisher={University of Chicago Press}
}

@inproceedings{grover2016node2vec,
  title={node2vec: Scalable feature learning for networks},
  author={Grover, Aditya and Leskovec, Jure},
  booktitle={Proceedings of the 22nd ACM SIGKDD international conference on Knowledge discovery and data mining},
  pages={855--864},
  year={2016}
}

@article{shchur2018pitfalls,
  title={Pitfalls of graph neural network evaluation},
  author={Shchur, Oleksandr and Mumme, Maximilian and Bojchevski, Aleksandar and G{\"u}nnemann, Stephan},
  journal={arXiv preprint arXiv:1811.05868},
  year={2018}
}

@article{kaufman2012leakage,
  title={Leakage in data mining: Formulation, detection, and avoidance},
  author={Kaufman, Shachar and Rosset, Saharon and Perlich, Claudia and Stitelman, Ori},
  journal={ACM Transactions on Knowledge Discovery from Data (TKDD)},
  volume={6},
  number={4},
  pages={1--21},
  year={2012},
  publisher={ACM New York, NY, USA}
}

@article{guan2023nba2vec,
  title={NBA2Vec: Dense feature representations of NBA players},
  author={Guan, Webster and Javed, Nauman and Lu, Peter},
  journal={arXiv preprint arXiv:2302.13386},
  year={2023}
}

@article{sun2019rotate,
  title={Rotate: Knowledge graph embedding by relational rotation in complex space},
  author={Sun, Zhiqing and Deng, Zhi-Hong and Nie, Jian-Yun and Tang, Jian},
  journal={arXiv preprint arXiv:1902.10197},
  year={2019}
}

@article{hamilton2017inductive,
  title={Inductive representation learning on large graphs},
  author={Hamilton, Will and Ying, Zhitao and Leskovec, Jure},
  journal={Advances in neural information processing systems},
  volume={30},
  year={2017}
}

@inproceedings{schlichtkrull2018modeling,
  title={Modeling relational data with graph convolutional networks},
  author={Schlichtkrull, Michael and Kipf, Thomas N and Bloem, Peter and Van Den Berg, Rianne and Titov, Ivan and Welling, Max},
  booktitle={European semantic web conference},
  pages={593--607},
  year={2018},
  organization={Springer}
}

@article{scully1974pay,
  title={Pay and performance in major league baseball},
  author={Scully, Gerald W},
  journal={The American Economic Review},
  volume={64},
  number={6},
  pages={915--930},
  year={1974},
  publisher={JSTOR}
}

@article{kahn1988racial,
  title={Racial differences in professional basketball players' compensation},
  author={Kahn, Lawrence M and Sherer, Peter D},
  journal={Journal of Labor Economics},
  volume={6},
  number={1},
  pages={40--61},
  year={1988},
  publisher={University of Chicago Press}
}

@article{berri1999most,
  title={Who is ‘most valuable’? Measuring the player's production of wins in the National Basketball Association},
  author={Berri, David J},
  journal={Managerial and decision economics},
  volume={20},
  number={8},
  pages={411--427},
  year={1999},
  publisher={Wiley Online Library}
}

@article{xu2025enhanced,
  title={Enhanced prediction of NBA players’ salaries using hybrid ensemble models and multi-objective optimization techniques based on the 2022--23 season dataset},
  author={Xu, Ziqi and Xu, Zitong},
  journal={Expert Systems with Applications},
  pages={128849},
  year={2025},
  publisher={Elsevier}
}
\end{document}